# Exploiting Adam-like Optimization Algorithms to Improve the Performance of Convolutional Neural Networks

Loris Nanni[1]*, Gianluca Maguolo[1] and Alessandra Lumini[2]
[1]DEI, University of Padua, viale Gradenigo 6, Padua, Italy
[2]DISI, Università di Bologna, Via dell'università 50, 47521 Cesena, Italy; alessandra.lumini@unibo.it
*Correspondence: loris.nanni@unipd.it;

**ABSTRACT**

Stochastic gradient descent (SGD) is the main approach for training deep networks: it moves towards the optimum of the cost function by iteratively updating the parameters of a model in the direction of the gradient of the loss evaluated on a minibatch. Several variants of SGD have been proposed to make adaptive step sizes for each parameter (adaptive gradient) and take into account the previous updates (momentum). Among several alternative of SGD the most popular are AdaGrad, AdaDelta, RMSProp and Adam which scale coordinates of the gradient by square roots of some form of averaging of the squared coordinates in the past gradients and automatically adjust the learning rate on a parameter basis. In this work, we compare Adam based variants based on the difference between the present and the past gradients, the step size is adjusted for each parameter. We run several tests benchmarking proposed methods using medical image data. The experiments are performed using ResNet50 architecture neural network. Moreover, we have tested ensemble of networks and the fusion with ResNet50 trained with stochastic gradient descent. To combine the set of ResNet50 the simple sum rule has been applied. Proposed ensemble obtains very high performance, it obtains accuracy comparable or better than actual state of the art. To improve reproducibility and research efficiency the MATLAB source code used for this research is available at GitHub: https://github.com/LorisNanni.

## 1. Introduction

Convolutional neural networks (CNNs) are the most effective tools in modern computer vision. In recent years they became the state of the art technique in many fields like image classification [1], object detection [2] and face recognition [3].

CNNs may include millions of parameters that must be learned by the minimization of a "loss function", i.e. by minimizing the deviation of predictions to the actual values of a training set. The minimization of the training loss of CNNs usually relies on algorithms based on gradient descent, the most naïve of which is the well-known stochastic gradient descent (SGD) [4]. However, SGD has some limitations since the optimization landscape is not convex and this might lead to the convergence to local minima. Hence, a lot of modifications of SGD have been proposed [5]–[10]. These modifications have two different goals: the first one is finding better minima of the training loss, trying not to get stacked in local minima; the second is to converge to a set of parameters that generalize well to a test set. These two goals are not necessarily correlated. A set of parameters might reach a very low training accuracy, while having a bad generalization to test data.

The first simple modification of the classical SGD method consists in exploiting its momentum to let the optimization process be guided also by its own inertia [5]. More or less, every modification of SGD consists in using the momentum or its past values to modify the current direction of the gradient. The same happens with AdaGrad [7] and its modification Adadelta [8]. The basic idea behind these two algorithms consists in decreasing the learning rate of variables with large partial derivatives more than the ones of variable with small derivatives. One of the most popular algorithm for training networks is Adam [9]. The idea behind Adam is to compute the moving average of the gradient and of its square. If the gradient changes often, in particular if it changes its sign often, the average of the square of the gradient will be large even if the average of the gradient is low. Using this information, Adam decreases the learning rate of those parameters whose gradient changes often.

Although Adam reaches very low minima of the training loss, experiments often shows that it does not obtain better results than SGD on the test set [11]. In [12], the training method switch from Adam to SGD to exploit the good convergence on the training set obtained by Adam and the generalization power of SGD. Modifications of Adam have been proposed aimed at reaching better generalization performances [10], [13], [14]. Nadam [13] introduces Nesterov momentum into Adam, while AMSGrad [14] ensure that the step size is non-increasing. Recently, diffGrad [10] proposed to adjust the step size for each parameter in order to make it proportional to its gradient changing; experiments reported in [10] showed that diffGrad reached state of the art results.

In this paper, we study the effect of the optimization process on the network performance: in our experiments we compare different optimization approaches in Convolutional Neural Networks (CNNs) for image classification tasks on multiple and diverse datasets. Selecting a fixed architecture (the well-known ResNet50 [1] in our experiments) and training the net by varying the optimization process, we obtain a large number of different networks that we use to create ensemble of CNNs.

At first, we introduce some Adam-based variants for deep network optimization then we compared them with SGD approach. The Adam based variants outperforms Adam, moreover, we show that changing the deep optimization is a feasible procedure for yields a set of different networks, providing a large number of partially independent classifiers. The classifiers can be fused together to create an ensemble of CNNs. Several works [15] have shown that ensembles of deep neural networks trained from random initialization are able to improve accuracy, uncertainty and out-of-distribution robustness of stand-alone deep learning models. In this work we exploit the instability of different optimization algorithms to design robust deep ensembles.

We evaluate our approach on three different datasets of medical image data, our experiments show that the proposed ensembles work well in all the tested problems gaining state-of-the-art classification performance.

To improve reproducibility and research efficiency the source code use for this research is available at GitHub https://github.com/LorisNanni.

## 2. Related Work

In this section we introduce with more details the optimization methods that we shall use in the rest of the paper.

### 2.1. Adam

Adam is an optimizer introduced in [9] which that computes adaptive learning rates for each parameter combining the ideas of momentum and adaptive gradient. Its update rule is based on the value of the gradient at the current step, and on the exponential moving averages of the gradient and its square. To be more precise, Adam defines the moving averages $m_t$ (the first moment) and $u_t$ (the second moment) as:

$$m_t = \rho_1 m_{t-1} + (1 - \rho_1) g_t \quad (1)$$

$$u_t = \rho_2 u_{t-1} + (1 - \rho_2) g_t^2 \quad (2)$$

where $g_t$ is the gradient at time $t$, the square on $g_t$ stands for the component-wise square, $\rho_1$ and $\rho_2$ are hyperparameters representing the exponential decay rate for the first moment and the second moment estimates (usually set to 0.9 and 0.999, respectively) and the moments are initialized to 0: $m_o = u_0 = 0$. In order to take into account the fact that, especially in the first steps, the value of moving averages will be very small due to their initialization to zero, the authors of Adam define a bias-corrected version of the moving averages:

$$\hat{m}_t = \frac{m_t}{(1 - \rho_1^t)} \quad (3)$$

$$\hat{u}_t = \frac{u_t}{(1 - \rho_2^t)} \quad (4)$$

The final update for each $\theta_t$ parameter of the network is:

$$\theta_t = \theta_{t-1} - \lambda \frac{\hat{m}_t}{\sqrt{\hat{u}_t} + \epsilon} \quad (5)$$

where $\lambda$ is the learning rate (usually set to 0.001), $\epsilon$ is a very small positive number to prevent any division by zero (usually set to $10^{-8}$) and all the operations are meant to be component-wise.

Notice that, while $g_t$ might have positive or negative components, $g_t^2$ has only positive components. Hence, if the gradient changes sign often, the value of $\hat{m}_t$ might be much lower than $\sqrt{\hat{u}_t}$. This means that in this case the step size is very small.

### 2.2. AMSGrad

AMSGrad [14] is a modification of Adam designed to deal with a convergence issue with Adam based optimizers. The AMSgrad formulation uses the maximum of past squared gradients to update the parameters rather than exponential average as in Adam. The definition of the first and second moments is the same used equations (1) and (2), but AMSGrad defines the non decreasing parameter $\bar{\bar{u}}_t$ as the maximum of past squared gradients:

$$\bar{\bar{u}}_t = \max(\bar{\bar{u}}_{t-1}, u_t) \quad (6)$$

The final update for each $\theta_t$ parameter of the network is similar to equation (5):

$$\theta_t = \theta_{t-1} - \lambda \frac{\hat{m}_t}{\sqrt{\bar{\bar{u}}_t} + \epsilon} \quad (7)$$

### 2.3. diffGrad

diffGrad is a method proposed in [10] that takes into account the difference of the gradient in order to set the learning rate. According to the observation that when gradient changes begin to reduce during training, this is often indicative of the presence of a global minima, diffGrad applies an adaptive adjusting driven by the difference between the present and the immediate past to lock parameters into global minima. Therefore, the step size is larger for faster gradient changing and a lower for lower gradient changing parameters. In order to define the update function, they define the absolute difference of two consecutive steps of the gradient as:

$$\Delta g_t = |g_{t-1} - g_t| \quad (8)$$

The final update for each $\theta_t$ parameter of the network is as in equation (5) where $\hat{m}_t, \hat{u}_t$ are defined as in equations (3) and (4) and the learning rate is modulated the Sigmoid of $\Delta g_t$ :

$$\xi_t = Sig(\Delta g_t) \quad (9)$$

$$\theta_{t+1} = \theta_t - \lambda \cdot \xi_t \frac{\hat{m}_t}{\sqrt{\hat{u}_t} + \epsilon} \quad (10)$$

where $Sig(\cdot)$ is the sigmoid function:

$$Sig(x) = \frac{1}{1 + e^{-x}} \quad (11)$$

## 3. Methods

In this work we compare several optimization methods for training CNN. For our experiments we selected as backbone architecture the ResNet50 [1], a variant of ResNet model composed by 48 Convolution layers, 1 MaxPool and 1 Average Pool layer. ResNet50 is one of the most widely used architecture, winner of ILSVRC 2015 contest. ResNet models are known for their skip connections, i.e. a feature transmission approach designed to prevent gradient vanishing, such that a much deeper network than those used previously could be effectively trained.

In this work we propose and evaluate the following variants of Adam optimization method:

- DGrad, it is a variant of original diffGrad, it is based on moving average of the element-wise squares of the parameter gradients;
- Cos#1 and Cos#2, are minor variant of DGrad based on the application of a cyclic learning rate [16] to DGrad;

The proposed approaches have different methods for defining $\xi_t$, then (10) is applied as in diffGrad.

**DGrad**
DGrad takes up the ideas of diffGrad defining the following absolute difference between two consecutive steps of the gradient:

$$\Delta ag_t = |g_t - avg_t| \quad (12)$$

where $avg_t$ contains the moving average of the element-wise squares of the parameter gradients, see eq. (2), we then normalize $\Delta ag_t$ by its maximum as

$$\Delta \widehat{ag_t} = \left(\frac{\Delta ag_t}{\max(\Delta ag_t)}\right) \quad (13)$$

and we define $\xi_t$ as:

$$\xi_t = Sig(4 \cdot \Delta \widehat{ag_t}) \quad (14)$$

The final update for each $\theta_t$ parameter of the network is as in equation (10). The rationale of the "4×" is to increase the range of the output of the sigmoid function.

**Cos#1 and Cos#2**
These two variants of DGrad exploits the idea of using a cyclic learning rate, which has proven to improve classification accuracy without a need to tune and often in fewer iterations [16].
We use the cos(x) periodic function to define a range of variation of the learning rate according to the following formulation:

$$lr_t = \left(2 - \left|\cos\left(\frac{\pi \cdot t}{steps}\right)\right| e^{-0.01 \cdot (mod(t,steps)+1)}\right) \quad (15)$$

Where *mod*() denote the function modulo and *steps*=30 is the period, where $lr_t$ is equal to 0 the value $9 \cdot 10^{-4}$ is assigned to it. The plot of $lr_t$ for t in the range 1:2×*steps* is reported in figure 1.

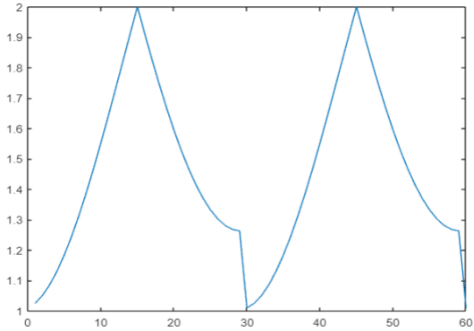

**Figure 1**. Cyclic learning rate.

In the method named **Cos#1** $lr_t$ is used as a multiplier of $\Delta \widehat{ag_t}$ in the definition of $\xi_t$, which becomes:

$$\xi_t = Sig(4 \cdot lr_t \cdot \Delta \widehat{ag_t}) \quad (16)$$

In the method named **Cos#2** the definition of $\xi_t$ includes a further additive factor $lra_t$ used to avoid a too small learning rate (as in $lr_t$ also for $lra_t$ we avoid zero numbers):

$$lra_t = \left(\left|\cos\left(\frac{\pi \cdot t}{steps}\right)\right| e^{-0.01 \cdot (mod(t,steps)+1)}\right) \quad (17)$$

$$\xi_t = Sig(2 \cdot lr_t \cdot \Delta \widehat{ag_t}) + Sig(4 \cdot lra_t) - 0.5 \quad (18)$$

In both methods the final update for each $\theta_t$ parameter of the network is performed as in equation (10).

All the CNNs have been trained using cross entropy as loss function and the following parameters:
- Batch size = 30;
- Number of epochs = 20;
- global learning rate = 0.001;

---
[1] https://ome.grc.nia.nih.gov/iicbu2008/hela/index.html
[2] https://zenodo.org/record/834910#.YFsmIa9KiCo

- gradient decay factor = 0.9;
- squared gradient decay factor = 0.999;
- loss function = cross entropy

Data augmentation is performed considering random reflection and random scale on both the axis.

**4. Experiments**

The following datasets have been used for assessing the performance of the tested approaches:
- HeLa, the 2D HELA dataset[1] [17] includes 832 grey-scale images of size 512×382 pixels divided into 10 classes (related to 10 different organelles). According to results reported in the literature we used a 5-fold cross validation testing protocol for this dataset.
- BG, the Breast Grading Carcinoma[2] [18] includes 300 RGB images of resolution 1280×960 pixels divided into 3 classes related to grades 1-3 of invasive ductal carcinoma of the breast. According to results reported in the literature we used a 5-fold cross validation testing protocol for this dataset.
- LAR, the Laryngeal data set[3] [19] includes 1320 patches of healthy and early-stage cancerous laryngeal tissues of size 100×100 equally divided into 4 classes: He (healthy tissue), Hbv (tissue with hypertrophic vessels), Le (tissue with leukoplakia) and IPCL (tissue with intrapapillary capillary loops). The dataset is already divided in 3 subfolders to be used for cross-validation purposes.

In the following table 1 we report the performance (average and standard deviation accuracy on 7 experiments) on the 3 datasets obtained by a stand-alone ResNet50 trained using different optimization methods:

- SGD: the original SGD approach is used;
- Adam: the original Adam approach is used;
- diffGrad: the net is trained according the diffGrad method [10];
- DGrad, Cos#1, Cos#2: the 3 variants proposed in section 3 are used for optimization.
- Res_SGD_x, fusion among x ResNet trained using SGD.

| accuracy | HeLa | | BG | | LAR | |
|---|---|---|---|---|---|---|
| | avg | std | avg | std | avg | std |
| SGD | **92.09** | 0.66 | 88.33 | 1.19 | 93.03 | 1.11 |
| Adam | 55.90 | 29.66 | 86.57 | 5.66 | 92.15 | 5.39 |
| diffGrad | 79.00 | 19.83 | 89.00 | 4.77 | 93.01 | 3.05 |
| DGrad | 75.25 | 22.56 | **89.29** | 4.21 | 91.07 | 3.79 |
| Cos#1 | 78.92 | 18.28 | 88.38 | 3.96 | 92.19 | 3.02 |
| Cos#2 | 66.25 | 28.74 | 88.05 | 6.36 | **93.04** | 2.99 |

**Table 1**. Performance of stand-alone CNN.

The performance reported in table 1 show that Adam methods obtain worse generalization than SGD, in the test data, as already shown in the literature. On average, the tested variants of Adam obtain better performance respect original Adam. Moreover, the high values of standard deviation for Adam and its variants reveal an high variability of result, which can be a drawback for a standalone approach but it is very appreciable for designing an ensemble of classifiers.

---
[3] https://zenodo.org/record/1003200#.YFsnR69KiCp

In table 2 we report the performance of deep ensembles obtained by the fusion with the sum rule of several CNNs. The number of classifiers included in the ensembles is enclosed in parentheses.

| accuracy | HeLa | BG | LAR |
|---|---|---|---|
| Adam(7) | 74.30 | 89.67 | **96.29** |
| diffGrad(7) | 94.88 | 91.67 | 95.91 |
| DGrad(7) | **95.35** | 92.67 | 94.85 |
| Cos#1(7) | 95.00 | **92.67** | 95.38 |
| Cos#2(7) | 91.05 | 92.00 | 95.98 |
| DGrad(7) + Cos#1(7) | 95.81 | 92.33 | 95.91 |
| DGrad(7) + Cos#1(7) + Cos#2(7) | 96.05 | **92.67** | 96.06 |
| DGrad(7) + Cos#1(7) + diffGrad(7) | **96.28** | 92.33 | 96.06 |
| DGrad(14) | 95.70 | **92.67** | 95.68 |
| DGrad(14) + Cos#1(7) | 95.58 | **92.67** | 96.29 |
| SGD(14) | **96.05** | **94.00** | 94.70 |
| SGD (7) | 95.70 | **94.00** | 94.32 |
| SGD(7) + DGrad(7) | 96.16 | 94.00 | 95.38 |
| SGD(14) + DGrad(7) + Cos#1(7) | 96.74 | **94.33** | 95.98 |
| SGD(14) + DGrad(7) + Cos#1(7) + diffGrad(7) | **96.98** | **94.33** | 96.14 |

**Table 2**. Performance of ensemble of CNN.

The following conclusion can be drawn from the table 2:
- The performance of Adam based approaches strongly improves considering ensemble of CNNs
- The performance of DGrad(14) is similar to that obtained by SGD(14);
- Adam variants, on average, perform better the original Adam;
- The fusion of CNNs trained by different optimization methods (i.e. SGD with Adam) permits to improve the performance: SGD(7) + DGrad(7) outperforms SGD(14) (ensembles with the same size). In our opinion, to combine networks trained using different optimization methods is a feasible way for building an ensemble.
- SGD(14) + DGrad(7) + Cos#1(7) obtains in the LAR dataset a F-measure of 95.99, SGD(14) + DGrad(7) + Cos#1(7) + diffGrad(7) obtains a F-measure of 96.15. Such result improve both the state-of-the-art performance reported in [19] (94.72) and in [20] (95.20) .

## 4. Conclusions

In this paper, we compared three different methods to modify the well know Adam procedure for deep network optimization. We obtain a set of very diverse classifiers and we show the usefulness to combine different CNNs for improving the performance. Finally, we compared and combined the proposed approach with CNNs trained using the widely used SGD.

The accuracy obtained by our ensemble in three different biomedical problems is higher than our baseline (the standard ResNet architecture trained by SGD) and outperforms each stand-alone method confirming that varying the optimization procedure is a feasible method to create deep ensembles. In this paper, we only used ResNet50 as backbone architecture, but we plan as a future work to make some experiments using lightweight CNN architectures. Deep ensembles require high computational power and long training/prediction time, but they permit a strong boost of performance.

**Acknowledgments:** We gratefully acknowledge the support of NVIDIA Corporation for the "NVIDIA Hardware Donation Grant" of a Titan X used in this research.